\documentclass{article}
\usepackage{spconf,amsmath,graphicx,setspace}

\usepackage{enumitem}
\setlist{nosep, leftmargin=14pt}

\usepackage{mwe} 

\usepackage{multirow}
\usepackage{booktabs} 
\usepackage{makecell}

\usepackage{url}
\usepackage[hidelinks]{hyperref}

\begin{document}

\title{Towards Cross-Domain Single Blood Cell Image Classification via Large-Scale LoRA-based Segment Anything Model}

\name{\begin{minipage}{0.95\textwidth}\centering Lingcong Cai$^{\star *}$ \qquad Yongcheng Li$^{\star *}$ \qquad Ying Lu$^{\S}$\sthanks{Lingcong Cai, Yongcheng Li, and Ying Lu contribute equally.} \qquad Yupeng Zhang$^{\P}$ \qquad Jingyan Jiang$^{\star}$\qquad Genan Dai$^{\star}$ \qquad Bowen Zhang$^{\star}$ \qquad Jingzhou Cao$^{\star}$ \qquad  Zhongxiang Zhang$^{\S \dagger}$ \qquad Xiaomao Fan$^{\star}$\sthanks{Xiaomao Fan and Zhongxiang Zhang are co-corresponding authors (Emails: astrofan@gmail.com, zhxzhong@mail.sysu.edu.cn).}\end{minipage}}


\address{$^{\star}$Shenzhen Technology University, Shenzhen, China \\
     $^{\S}$The Third Affiliated Hospital of Sun Yat-sen University, Guangzhou, China \\
     $^{\P}$South China Normal University, Guangzhou, China}

\maketitle

\begin{abstract}
Accurate classification of blood cells plays a vital role in hematological analysis as it aids physicians in diagnosing various medical conditions. In this study, we present a novel approach for classifying blood cell images known as BC-SAM. BC-SAM leverages the large-scale foundation model of Segment Anything Model (SAM) and incorporates a fine-tuning technique using LoRA, allowing it to extract general image embeddings from blood cell images. To enhance the applicability of BC-SAM across different blood cell image datasets, we introduce an unsupervised cross-domain autoencoder that focuses on learning intrinsic features while suppressing artifacts in the images. To assess the performance of BC-SAM, we employ four widely used machine learning classifiers (Random Forest, Support Vector Machine, Artificial Neural Network, and XGBoost) to construct blood cell classification models and compare them against existing state-of-the-art methods. Experimental results conducted on two publicly available blood cell datasets (Matek-19 and Acevedo-20) demonstrate that our proposed BC-SAM achieves a new state-of-the-art result, surpassing the baseline methods with a significant improvement. The source code of this paper is available at \href{https://github.com/AnoK3111/BC-SAM}{https://github.com/AnoK3111/BC-SAM}.
\end{abstract}

\section{Introduction}
\label{sec:intro}
Blood cell morphology detection is a common method for disease diagnosis. It assists physicians in the diagnosis and monitoring of various medical conditions, including leukemia, anemia, infections, autoimmune disorders, and other blood-related diseases\cite{quan2020effective,anilkumar2020survey}. Therefore, accurately and rapidly classifying blood cells is of significant importance for the timely detection of blood cell diseases and for the prompt treatment of patients.

In clinical practice, the identification of blood cell morphology is traditionally carried out manually by trained experts using a microscope, which is a time-consuming and labor-intensive process. However, with the remarkable advancements of deep learning technologies in the field of computer vision, researchers have endeavored to automate blood cell segmentation and identification using deep neural network models, demonstrating the immense potential for blood cell detection \cite{xie2017aggregated,acevedo2021new}. Xie et al. \cite{xie2017aggregated} proposed a highly accurate approach based on the ResNext architecture for blood cell identification in blood smears, while Acevedo et al. \cite{acevedo2021new} presented a predictive model for automatically identifying patients with myelodysplastic syndrome. These studies showcased impressive performance on specific datasets. Nonetheless, when applied to unseen datasets, their performance deteriorated rapidly due to a shift in data distribution, commonly referred to as domain shift. This shift is caused by various external factors, such as differences in microscope settings, choice of staining solution, and lighting intensity. To address the challenge posed by domain-specific information, Salehi et al. \cite{salehi2022unsupervised} proposed an unsupervised cross-domain feature learning model for classifying blood cell images collected from different sources. Although their approach yielded improved performance within the target domain, there is still significant room for enhancing its generalizability.

Recently, significant advancements have been made in large-scale foundation models for natural language processing tasks, such as LLaMA \cite{touvron2023llama} and GPT-4 \cite{nori2023capabilities}, as well as in computer vision tasks, including SAM \cite{kirillov2023segment} and SegGPT \cite{wang2023seggpt}. Researchers have also ventured into developing large-scale foundation models specific to the medical domain, such as Med-SAM\cite{ma2023segment}, and SAMMed-3D\cite{wang2023sam}. Drawing inspiration from the remarkable versatility of foundation models, we employ the SAM model as a foundation for learning image embeddings from blood cell images with the aid of LoRA \cite{hu2021lora} for fine-tuning. It is important to note that we select SAM as our baseline for image segmentation due to the close resemblance between blood cell images and natural images. To further enhance the model's performance across different domains, we introduce a cross-domain autoencoder that focuses on learning intrinsic information while suppressing artifacts in the blood cell images. Finally, we construct blood cell identification models using four widely used machine learning algorithms: Random Forest (RF), Support Vector Machine (SVM), Artificial Neural Network (ANN), and XGBoost. Experimental results on two domain datasets, Matek-19 and Acevedo-20, demonstrate that our proposed BC-SAM achieves state-of-the-art cross-domain performance by a significant margin. In summary, the primary contributions of our work can be summarized as follows:

\begin{itemize}
    \item We first introduce the large-scale foundation model SAM into the blood cell identification task to learn general image embeddings, leveraging LoRA for fine-tuning.
    
    \item We propose a novel framework of combining LoRA-based SAM and cross-domain autoencoder to learn rich intrinsic features from blood cell images while effectively muting the domain-specific artifact information.  
    
    \item Experiment results show that our proposed BC-SAM can achieve a new cross-domain state-of-the-art result in blood cell identification.
\end{itemize}

\section{MATERIALS AND METHODS}
\label{sec:materials}
Fig.\ref{fig:pipeline} shows the pipeline of blood cell classification, which is described in detail as follows:

\begin{figure}[tb]
\centering
\includegraphics[width=1.0\linewidth]{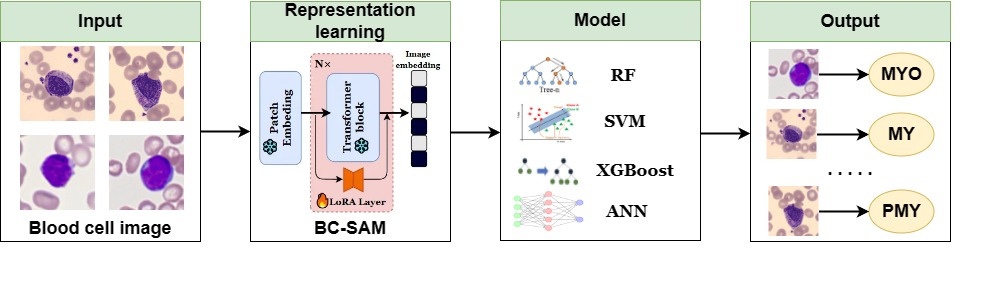}
\caption{The pipeline of blood cell classification. It mainly consists of image representation learning with a segment anything model (SAM) and classification model building with four widely used machine learning methods.}
\label{fig:pipeline}
\end{figure}

\subsection{Datasets}
In this section, we utilize two different domain blood cell datasets of Make-19 and Acevedo-20 to evaluate the BC-SAM classification performance. The details are described as follows:

\textbf{Matek-19}\cite{matek2019human}: it comprised more than 18,000 annotated blood cell images with a total of 15 distinct classes. The images have dimensions of \(400 \times 400\) pixels, which correspond to approximately \(29 \times 29\) micrometers in size.

\textbf{Acevedo-20}\cite{acevedo2020dataset}: it consisted of more than 17,000 individual normal cell images with 8 classes in total. The images in this dataset have dimensions of \(360 \times 363\) pixels, equivalent to \(36 \times 36.3\) micrometers in size.

\begin{figure}[tb]
\centering
	\includegraphics[width=1.0\linewidth]{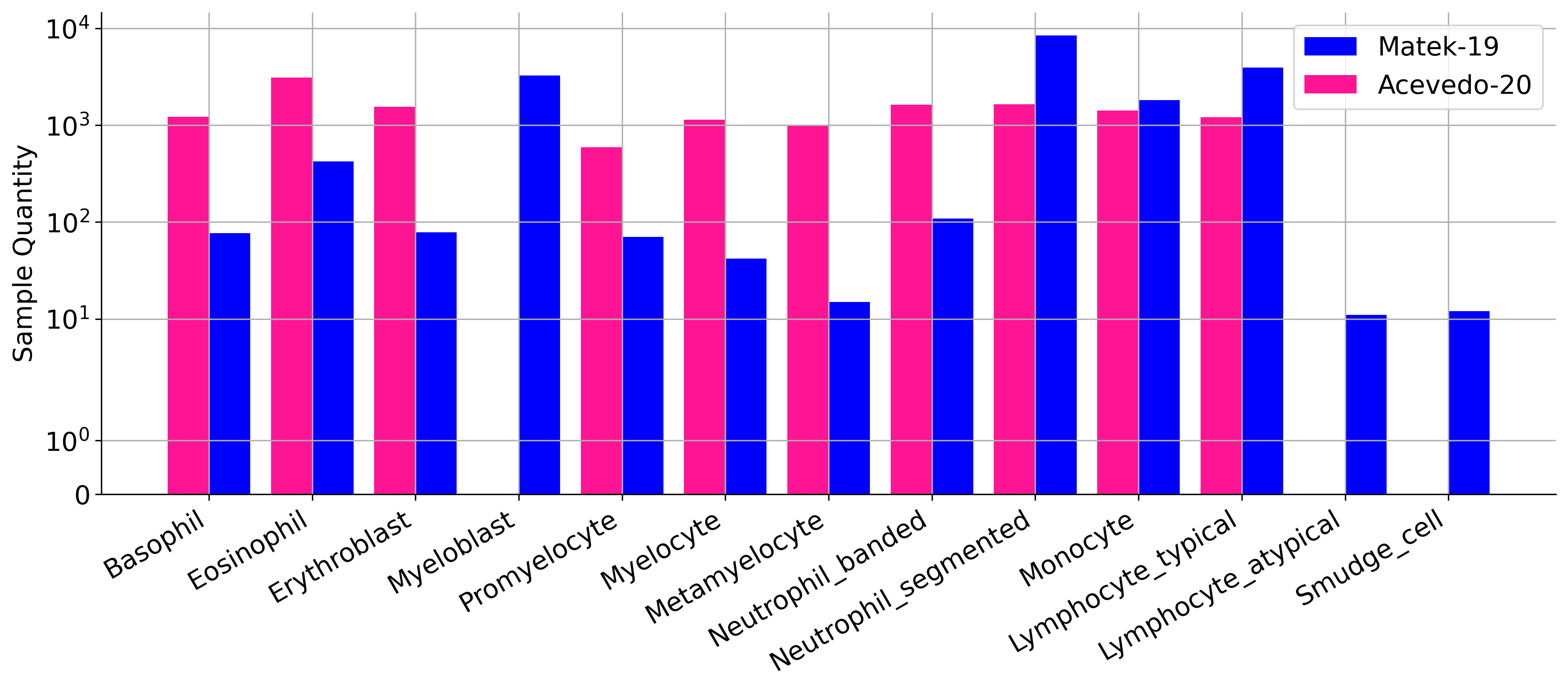}
\caption{The category distribution of blood cell image datasets.}
\label{fig:dataset}
\end{figure}

 As shown in Fig.\ref{fig:dataset}, we categorize the blood cell images into 13 classes based on their labels. All blood cell images are resized to \(224 \times 224\) in our experiments.

\subsection{Image representation learning}
\begin{figure}[tb]
\centering
	\includegraphics[width=1\linewidth]{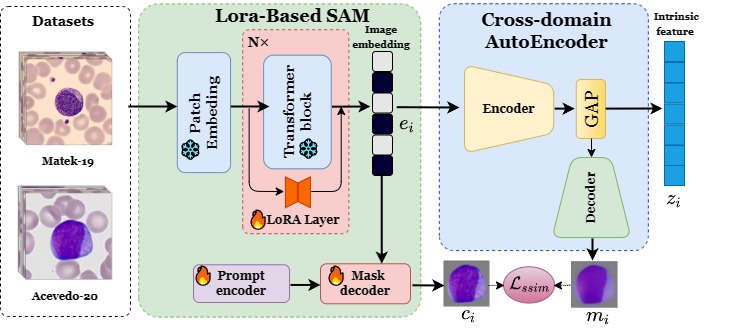}
\caption{The architecture of BC-SAM. It consists of two components: LoRA-based segment anything model and cross-domain autoencoder. Snowflake refers to the frozen parameters while flame refers to the learnable parameters.}
\label{fig:bc-sam}
\end{figure}

In this section, we introduce a new framework called the Blood Cell Segment Anything Model (BC-SAM) to extract rich intrinsic features from blood cell images while suppressing domain-specific artifacts. The architecture of BC-SAM, as illustrated in Fig. \ref{fig:bc-sam}, consists of two main components: the LoRA-Based Segment Anything Model (LoRA-SAM) and the Cross-Domain Autoencoder. Initially, we train LoRA-SAM to learn image embeddings and perform image segmentation. Subsequently, the cross-domain autoencoder is trained to extract intrinsic features.

\subsubsection{LoRA-based segment anything model}
In the central portion of Fig. \ref{fig:bc-sam}, we introduce a LoRA-based SAM approach to learn the image embedding from a blood cell image and generate the corresponding segmentation mask using the default prompt encoder. It is important to note that SAM possesses impressive zero-shot generalization capabilities when dealing with unseen images. However, its performance experiences significant degradation when directly applied to a new domain-specific image segmentation task, even with the utilization of various prompt schemes. To address this issue, fine-tuning the parameters of SAM enables the model to effectively learn domain-specific feature information \cite{xiong2023mammo,fazekas2023adapting}. Instead of fine-tuning the entire set of SAM parameters, which requires limited domain-specific data and is computationally inefficient, we adopt a lightweight parameter fine-tuning technique known as LoRA (Hu et al., 2021). LoRA provides a resource-efficient and precise fine-tuning approach specifically designed for large-scale foundation models. Specifically, we freeze the transformer blocks in SAM to keep the parameters $W$ fixed and introduce trainable LoRA Layers $\Delta W$. The structure of the LoRA layer comprises two linear layers, denoted as \(A \in {R^{r \times {C_{in}}}}\) and \(B \in {R^{{C_{out}} \times r}}\), where \(r \ll \min \{ {C_{in}},{C_{out}}\}\). Here, $r$ represents the intrinsic dimension and is set to 4. Consequently, the parameters of the update layers can be defined as follows:

\begin{equation}
\hat{W} = W + \Delta W = W + BA
\end{equation}

The prompt encoder, along with the subsequent mask decoder, is composed of lightweight transformer layers, allowing for direct fine-tuning without the need for LoRA layers. Leveraging SAM's impressive few-shot adaptation capability, we achieve strong segmentation performance using only 1 \% of blood cell images with masked annotations. The mask decoder ultimately produces cell segmentation masks. In order to learn intrinsic features from cross-domain datasets, a post-processing step is performed in Section \ref{sec:cd-autoencoder} on the original blood cell image after obtaining the segmentation masks. This involves cropping the original image to retain only the square regions corresponding to the masked areas, with the grayscale outside the masked area set to 128. Formally, given a image $X_{i}$ from the $j$-th domain dataset $D_{j}$, we have

\begin{equation}
    e_{i}, c_{i} = f_{LoRA-SAM}(X_{i};A, B)
\end{equation}

\noindent where $e_{i}$ is the learned image embedding from LoRA-based SAM $f_{LoRA-SAM}(\cdot)$ and $c_{i}$ is the post-processed single blood cell image.

\begin{table*}[!ht]

\belowrulesep=0pt
\aboverulesep=0pt
\tabcolsep=0.4cm
\center
\caption{The accuracy score of different classification models tested on source and target domain datasets}
\begin{tabular}{c c c c c c}
\label{classification result}
Trained on& \multicolumn{2}{c}{Matek-19} & \multicolumn{2}{c}{Acevedo-20} & \multirow{3}{4.5em}{Average(\%)}\\ 
\cmidrule(lr){0-0}\cmidrule(lr){2-3}\cmidrule(lr){4-5}
Tested on& Matek-19(\%) & Acevedo-20(\%) & Matek-19(\%) & Acevedo-20(\%)\\ \specialrule{0em}{1pt}{0pt}\hline
\specialrule{0em}{2pt}{0pt}
ResNext\cite{matek2019human} &\textbf{96.10} &8.10 &7.30$\pm$3.1 & \textbf{85.70$\pm$2.4} & 49.30\\
AE-CFE-RF\cite{salehi2022unsupervised} &83.70$\pm$0.5 &21.90$\pm$0.4 &45.10$\pm$0.5 & 65.20$\pm$0.5 & 53.98\\
\hline
\specialrule{0em}{2pt}{0pt}
BC-SAM-RF & 89.03$\pm$0.4 & 24.07$\pm$0.4 & \textbf{51.94$\pm$0.8} & 70.41$\pm$1.0 & 58.86\\ 
BC-SAM-SVM(poly) & 90.95$\pm$0.4 & 24.77$\pm$0.6 & 50.67$\pm$0.4 & 72.99$\pm$0.4 & 59.85\\ 
BC-SAM-XGBoost & 91.41$\pm$0.5 & 36.84$\pm$1.0 & 49.22$\pm$1.0 & 74.52$\pm$0.7  & 63.00\\ 
BC-SAM-SVM(rbf) & 92.51$\pm$0.4 & \textbf{47.50$\pm$0.6} & 34.53$\pm$1.3 & 78.27$\pm$0.5 & 63.20\\
BC-SAM-ANN & 92.28$\pm$0.3 & 41.21$\pm$1.0 & 42.93$\pm$2.4 & 77.74$\pm$0.9 & \textbf{63.54}\\
\hline
\end{tabular}
\end{table*}

\subsubsection{Cross-domain autoencoder}
\label{sec:cd-autoencoder}
The cross-domain autoencoder comprises two main components: the encoder and the decoder. The encoder is a lightweight network that includes three convolution layers and a global average pooling (GAP) layer. This design effectively addresses the overfitting issue and accelerates the training process of the model. Our encoder, denoted as $f_{enc}(\cdot)$, maps the image embedding $e_i$ from LoRA-SAM to a 50-dimensional latent representation $z_i$, which represents the intrinsic features of the blood cell image. The decoder consists of eight transposed convolution layers. To learn the intrinsic features $z_i$ from the blood cell image and reduce artifacts in the image background, our decoder focuses solely on reconstructing the post-processed image mentioned earlier. Thus, our proposed cross-domain autoencoder can be formulated as:

\begin{eqnarray}
    z_{i}&=&f_{enc}(e_{i};\theta_{e}) \\
    m_{i}&=&f_{dec}(z_{i};\theta_{d})
\end{eqnarray}

\noindent where $m_{i}$ represents the image reconstructed by the decoder, $\theta_{e}$ and $\theta_{d}$ are the parameters of the encoder and the decoder. 

To mute the domain-specific artifact information and learn the intrinsic feature information, we utilize the Structure Similarity Index Measure(SSIM)\cite{1284395} to maximize the similarity between the reconstructed image $m_{i}$ by cross-domain autoencoder and post-processed image $c_{i}$ generated by LoRA-SAM. In other words, the similarity loss $\mathcal{L}_{ssim}$ can be formally defined as:

\begin{equation}
\label{ssim}
\mathcal{L}_{ssim}=1-SSIM(m_{i}, c_{i})
\end{equation}


\noindent $SSIM(\cdot, \cdot)$ is the function of Structure Similarity Index Measure. What's more, to align the latent representation obtained from different blood cell datasets, we use Maximum Mean Discrepancy(MMD)\cite{tzeng2014deep} $\mathcal{L}_{mmd}$. Therefore, the total objective loss $L$ of BC-SAM can be defined to be:

\begin{equation}
\mathcal{L} = \mathcal{L}_{ssim} + \lambda\mathcal{L}_{mmd}
\end{equation}

\noindent where $\lambda$ is a trade-off coefficient, which is set to 5 in this work.

\subsection{Blood cell classifier}
In this section, using the learned intrinsic features $z_i$ by BC-SAM as input, we utilize four widely used machine learning methods of RF, SVM, XGBoost and ANN to build blood cell classification models. On the other side, these classification models can also further verify the cross-domain feature learning capability of our proposed BC-SAM. In this study, we employ the accuracy score to evaluate the blood cell identification performance.

\section{RESULTS AND DISCUSSION}
\subsection{Implementation details}
All the experiments are conducted on a computing server equipped with an NVIDIA RTX6000 GPU offering 24GB of video memory and dual Intel(R) Xeon(R) Gold 6248R CPUs providing 256GB of system memory. The installed operating system is Ubuntu 20.04.5 LTS. BC-SAM and the blood cell classifiers are implemented using Pytorch 1.9.1 and Scikit-learn 1.3.1.

In this study, we employ the AdamW to optimize BC-SAM with an initial learning rate of 0.0005 and a weight decay rate of 0.05. In addition, warm-up and cosine learning schedules are presented to dynamically control the learning rate. LoRA-SAM is trained for 85 epochs while the cross-domain autoencoder is trained for 10 epochs.

As for the blood cell classifier, we utilize four machine learning classifiers, their hyperparameters are set as follows: 1)RF: 200 estimators and the maximum depth is 16; 2)SVM: RBF and Poly kernels; 3)ANN: one hidden layer with size of 100; 4)XGBoost: default.

\subsection{Blood cell classifying performance}
To evaluate the performance of our proposed framework, we utilize the 5-fold cross-validation to verify the model performance on single and cross-domain datatsets. Specifically, as for the performance on a single dataset, we utilize the 5-fold cross-validation to compute an averaging accuracy score to evaluate the proposed framework performance. Regarding the performance on a cross-domain dataset, we train the model on a single dataset with cross-domain validation. For each fold cross-validation, the trained model is tested on an external dataset. Finally, computing the each-fold average accuracy scores on the external dataset is the cross-domain performance.

Table \ref{classification result} shows the accuracy scores of different classification models on source and target domain datasets. It is observed that BC-SAM combining SVM with RBF kernel (\emph{i.e.}, BC-SAM-SVM(rbf)) can achieve quite competitive performance on source domain dataset, which is up to 92.51 \% and 78.27 \% accuracy scores on Matek-19 and Acevedo-20 datasets, respectively. Regarding the cross-domain performance, BC-SAM combing SVM with RBF kernel achieves the best cross-domain accuracy scores on Matek-19 and Acevedo-20 datasets. Specifically, BC-SAM-SVM(rbf) trained on Matek-19 while tested on Acevedo-20 obtains 47.50 \% accuracy score while BC-SAM-RF trained on Acevedo-20 while tested on Matek-19 obtains 51.94 \% accuracy score. In addition, we calculate the average accuracy scores among source and target domain datasets. We can see that BC-SAM combining ANN (\emph{i.e.}, BC-SAM-ANN) achieves the best comprehensive accuracy score of 63.54 \%.

Compared with extant baselines, ResNext \cite{matek2019human} can achieve the best accuracy scores on source domain datasets. In this work, ResNext merely optimizes the model on a single dataset, leading to quite poor performance on target domain performance. Regarding AE-CFE-RF \cite{salehi2022unsupervised}, it is a cross-domain adapted model to extract features in an unsupervised manner on different-domain datasets. AE-CFE-RF achieved a big improvement in target domain performance. As shown in Table \ref{classification result}, our proposed framework with different machine learning methods is superior to AE-CFE-RF on both source domain and target domain accuracy scores, with over 10 \% average accuracy score. It means that our proposed BC-SAM based on a large-scale foundation model SAM has the powerful capability to learn the intrinsic feature information from blood cell images.

\section{Conclusion}
\label{conclusion}
In this study, we first propose a novel blood cell identification framework called BC-SAM based large-scale LoRA-based segment anything model. Specifically, it mainly consists of two components: Lora-based SAM and cross-domain autoencoder. As for the Lora-based SAM, we utilize SAM with LoRA for fine-tuning domain-specific parameters to learn image embedding and blood cell segmentation masks. Regarding the cross-domain autoencoder, we present a cross-domain autoencoder with maximizes the similarity between the post-processed image and reconstructed blood cell image. By this, BC-SAM can focus on learning intrinsic information and suppressing artifacts in the blood cell images. To evaluate the blood cell identification performance, four widely used machine learning methods are utilized as classifiers. Experiment results demonstrate that our proposed BC-SAM achieves a new state-of-the-art baseline, which surpasses counterparts on cross-domain accuracy by a big margin. 

\section{Acknowledgement}
This work is partially supported by Special subject on Agriculture and Social Development, Key Research and Development Plan in Guangzhou (2023B03J0172), Basic and Applied Basic Research Project of Guangdong Province (2022B1515130009), Natural Science Foundation of Top Talent of SZTU (GDRC202318), Stable Support Project for Shenzhen Higher Education Institutions (SZWD2021011), and Research Promotion Project of Key Construction Discipline in Guangdong Province (2022ZDJS112).


\begin{thebibliography}{10}

\bibitem{quan2020effective}
Quan Quan, Jianxin Wang, and Liangliang Liu,
\newblock ``An effective convolutional neural network for classifying red blood cells in malaria diseases,''
\newblock {\em Interdisciplinary Sciences: Computational Life Sciences}, vol. 12, pp. 217--225, 2020.

\bibitem{anilkumar2020survey}
KK~Anilkumar, VJ~Manoj, and TM~Sagi,
\newblock ``A survey on image segmentation of blood and bone marrow smear images with emphasis to automated detection of leukemia,''
\newblock {\em Biocybernetics and Biomedical Engineering}, vol. 40, no. 4, pp. 1406--1420, 2020.

\bibitem{xie2017aggregated}
Saining Xie, Ross Girshick, Piotr Doll{\'a}r, Zhuowen Tu, and Kaiming He,
\newblock ``Aggregated residual transformations for deep neural networks,''
\newblock in {\em Proceedings of the IEEE conference on computer vision and pattern recognition}, 2017, pp. 1492--1500.

\bibitem{acevedo2021new}
Andrea Acevedo, Anna Merino, Laura Bold{\'u}, Angel Molina, Santiago Alf{\'e}rez, and Jos{\'e} Rodellar,
\newblock ``A new convolutional neural network predictive model for the automatic recognition of hypogranulated neutrophils in myelodysplastic syndromes,''
\newblock {\em Computers in Biology and Medicine}, vol. 134, pp. 104479, 2021.

\bibitem{salehi2022unsupervised}
Raheleh Salehi, Ario Sadafi, Armin Gruber, Peter Lienemann, Nassir Navab, Shadi Albarqouni, and Carsten Marr,
\newblock ``Unsupervised cross-domain feature extraction for single blood cell image classification,''
\newblock in {\em International Conference on Medical Image Computing and Computer-Assisted Intervention}. Springer, 2022, pp. 739--748.

\bibitem{touvron2023llama}
Hugo Touvron, Thibaut Lavril, Gautier Izacard, Xavier Martinet, Marie-Anne Lachaux, Timoth{\'e}e Lacroix, Baptiste Rozi{\`e}re, Naman Goyal, Eric Hambro, Faisal Azhar, et~al.,
\newblock ``Llama: Open and efficient foundation language models,''
\newblock {\em arXiv preprint arXiv:2302.13971}, 2023.

\bibitem{nori2023capabilities}
Harsha Nori, Nicholas King, Scott~Mayer McKinney, Dean Carignan, and Eric Horvitz,
\newblock ``Capabilities of gpt-4 on medical challenge problems,''
\newblock {\em arXiv preprint arXiv:2303.13375}, 2023.

\bibitem{kirillov2023segment}
Alexander Kirillov, Eric Mintun, Nikhila Ravi, Hanzi Mao, Chloe Rolland, Laura Gustafson, Tete Xiao, Spencer Whitehead, Alexander~C Berg, Wan-Yen Lo, et~al.,
\newblock ``Segment anything,''
\newblock {\em arXiv preprint arXiv:2304.02643}, 2023.

\bibitem{wang2023seggpt}
Xinlong Wang, Xiaosong Zhang, Yue Cao, Wen Wang, Chunhua Shen, and Tiejun Huang,
\newblock ``Seggpt: Segmenting everything in context,''
\newblock {\em arXiv preprint arXiv:2304.03284}, 2023.

\bibitem{ma2023segment}
Jun Ma and Bo~Wang,
\newblock ``Segment anything in medical images,''
\newblock {\em arXiv preprint arXiv:2304.12306}, 2023.

\bibitem{wang2023sam}
Haoyu Wang, Sizheng Guo, Jin Ye, Zhongying Deng, Junlong Cheng, Tianbin Li, Jianpin Chen, Yanzhou Su, Ziyan Huang, Yiqing Shen, et~al.,
\newblock ``Sam-med3d,''
\newblock {\em arXiv preprint arXiv:2310.15161}, 2023.

\bibitem{hu2021lora}
Edward~J Hu, Yelong Shen, Phillip Wallis, Zeyuan Allen-Zhu, Yuanzhi Li, Shean Wang, Lu~Wang, and Weizhu Chen,
\newblock ``Lora: Low-rank adaptation of large language models,''
\newblock {\em arXiv preprint arXiv:2106.09685}, 2021.

\bibitem{matek2019human}
Christian Matek, Simone Schwarz, Karsten Spiekermann, and Carsten Marr,
\newblock ``Human-level recognition of blast cells in acute myeloid leukaemia with convolutional neural networks,''
\newblock {\em Nature Machine Intelligence}, vol. 1, no. 11, pp. 538--544, 2019.

\bibitem{acevedo2020dataset}
Andrea Acevedo, Anna Merino, Santiago Alf{\'e}rez, {\'A}ngel Molina, Laura Bold{\'u}, and Jos{\'e} Rodellar,
\newblock ``A dataset of microscopic peripheral blood cell images for development of automatic recognition systems,''
\newblock {\em Data in brief}, vol. 30, 2020.

\bibitem{xiong2023mammo}
Xinyu Xiong, Churan Wang, Wenxue Li, and Guanbin Li,
\newblock ``Mammo-sam: Adapting foundation segment anything model for automatic breast mass segmentation in whole mammograms,''
\newblock in {\em International Workshop on Machine Learning in Medical Imaging}. Springer, 2023, pp. 176--185.

\bibitem{fazekas2023adapting}
Botond Fazekas, Jos{\'e} Morano, Dmitrii Lachinov, Guilherme Aresta, and Hrvoje Bogunovi{\'c},
\newblock ``Adapting segment anything model (sam) for retinal oct,''
\newblock in {\em International Workshop on Ophthalmic Medical Image Analysis}. Springer, 2023, pp. 92--101.

\bibitem{1284395}
Zhou Wang, A.C. Bovik, H.R. Sheikh, and E.P. Simoncelli,
\newblock ``Image quality assessment: from error visibility to structural similarity,''
\newblock {\em IEEE Transactions on Image Processing}, vol. 13, no. 4, pp. 600--612, 2004.

\bibitem{tzeng2014deep}
Eric Tzeng, Judy Hoffman, Ning Zhang, Kate Saenko, and Trevor Darrell,
\newblock ``Deep domain confusion: Maximizing for domain invariance,''
\newblock {\em arXiv preprint arXiv:1412.3474}, 2014.

\end{thebibliography}

\end{document}